# Robustness Enhancement of Object Detection in Advanced Driver Assistance Systems (ADAS)

Le-Anh Tran
*Dept. of Electronics Engineering*
*Myongji University*
Gyeonggi, South Korea
leanhtran@mju.ac.kr

Truong-Dong Do
*Dept. of Artificial Intelligence Convergence*
*Chonnam University*
Gwangju, South Korea
truongdong.cnu@gmail.com

Dong-Chul Park
*Dept. of Electronics Engineering*
*Myongji University*
Gyeonggi, South Korea
parkd@mju.ac.kr

My-Ha Le
*Faculty of Electrical and Electrinics Engineering*
*HCMC University of Technology and Education*
Ho Chi Minh City, Vietnam
halm@hcmute.edu.vn

*Abstract* — A unified system integrating a compact object detector and a surrounding environmental condition classifier for enhancing the robustness of object detection scheme in advanced driver assistance systems (ADAS) is proposed in this paper. ADAS are invented to improve traffic safety and effectiveness in autonomous driving systems where the object detection plays an extremely important role. However, modern object detectors integrated in ADAS are still unstable due to high latency and the variation of the environmental contexts in the deployment phase. Our system is proposed to address the aforementioned problems. The proposed system includes two main components: (1) a compact one-stage object detector which is expected to be able to perform at a comparable accuracy compared to state-of-the-art object detectors, and (2) an environmental condition detector that helps to send a warning signal to the cloud in case the self-driving car needs human actions due to the significance of the situation. The empirical results prove the reliability and the scalability of the proposed system to realistic scenarios.

## I. INTRODUCTION

Recent technological breakthroughs of convolution neural networks (CNNs) and the outstanding evolution of Graphics Processing Units (GPUs) that can boost the performance of parallel computation have made deep learning become the dominating approach for various computer vision tasks. CNN-based object detection, particularly, is a computer technology that has attracted a huge party of researchers for the past decade because of its applicability. Generally, there exist two genres of object detectors: one-stage object detectors which show high inference speed and considerable accuracy, the most popular one for this one-stage type is YOLO [1,2], because of the superiority in inference speed, one-stage object detectors generally are integrated into many real-time object detection systems and mobile devices; two-stage object detectors, alternatively, show higher object recognition and localization accuracy but with more expensive computational cost and diminished speed such as Faster R-CNN [3].

Autonomous driving vehicles, on the other hand, have been considered the future of technology as they have drawn huge attention for the last decade. Many studies regarding autonomous robots are conducted such as autonomous drones [4] and self-driving cars [5-7]. Advanced driver assistance systems (ADAS), consequently, are introduced to improve traffic effectiveness, prevent traffic accidents, and facilitate fully autonomous driving in near future. However, modern object detectors deployed in ADAS are still unstable due to many factors including how to select an appropriate network architecture that can properly balance the speed-accuracy trade-off. The inference speed is vastly important because deployment to mobile devices has many stringent requirements on latency and computational resources, while mobile devices usually support limited built-in hardware resources. We are in need of reducing the latency by using relatively small networks while maintaining the accuracy as much as possible.

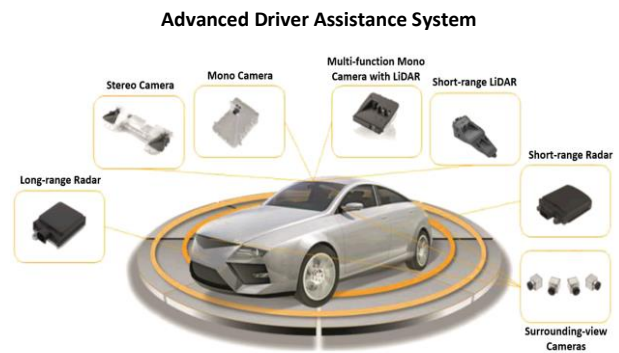

Fig. 1. Advanced driver assistance systems (ADAS) [8].

The goal of this paper is to enhance the robustness of the object detection module in ADAS by determining an object detection network that is able to balance efficiently the trade-off between inference speed and detection accuracy. In this paper, we propose a YOLO-based object detector which is constructed based on YOLOv2 [1] with only 17 convolutional layers in its backbone in order to achieve a low latency as well as a favorable detection accuracy. We also find that a critical reason for the instability of object detectors during the deployment phase is the variation of the environmental contexts. An object detector is trained with certain weather scenes may not perform properly with other weather scenes that it is not trained to work with. To address this problem, we further propose an environmental condition classifier and a communication protocol between the system and the cloud via internet connection so that the system is able to send a warning signal to the cloud in case the self-driving car needs human actions due to the significance of the situation.

We first briefly introduce ADAS and their challenges in Section II. Then, in Section III, we propose our system that can be used for improving the robustness of the object detection module in ADAS. The experiments are discussed in Section IV. At last, conclusions of the paper and the future works are provided in Section V.

## II. ADAS

### A. Brief Introduction to ADAS

Advanced driver assistance systems (ADAS) are automated electronic systems which assist drivers in driving and parking in order to increase car and road safety. ADAS use a set of sensors and cameras such as ultrasonic, radar, LiDAR, infrared and vision sensors, 2D camera, and 3D camera to detect nearby obstacles or driver errors and predict possible dangers on an earlier stage then respond to dangerous situations accordingly by analyzing the behavior of the vehicles Due to the increasing need for mobility, traffic has become more complex and therefore has become a greater challenge for all vehicle users. Hence, ADAS are crucial in order to avoid accidents and possible injuries or fatalities [8]. Though ADAS provide solutions for intelligent driving, they do not act autonomously but simply provide additional information about the traffic situation to support and assist drivers in implementing essential actions. The synchronization of the actions of the driver and the information from the environment and furthermore the recognition of the current situation is extremely essential for obtaining the best of ADAS. For the last few years, various types of complex control units have been developed and integrated into ADAS [8]. An example of ADAS is illustrated in Fig. 1 which is adapted from [8].

### B. Challenges and Proposals

As mentioned earlier, ADAS gather information from the surrounding environment to support driving, and the object detection plays an extremely crucial role in ADAS. Object detectors in ADAS currently have faced many challenges due to the demands on inference speed and accuracy. While the inference speed depends mainly on hardware resources and the complexity of the network, the accuracy, on the other hand, depends completely on the detection algorithm adopted for the system. In the scenario that the hardware technology is developing rapidly, the future detection algorithms are expected to perform at higher speeds in near future. Currently, the best choice for the object detector in a self-driving car is a network which is able to balance efficiently the inference speed and detection accuracy. In order to achieve this purpose, we propose a YOLO-based object detector that can ensure a low latency as well as a favorable detection accuracy.

In the driving scenarios, object detection models popularly are trained and tested on driving datasets such as BDD100K [9] and Cityscapes [10]. However, those datasets include image data that are collected in certain areas, which means that an object detector which is trained on mentioned datasets may not perform as designed for when it is tested with another dataset which contains image data gathered from another city with very different scene contexts. The problem can be seen from a simpler perspective when an autonomous car operates in many different weather conditions in a day or different light conditions between daytime and nighttime; for instance, the object detector of an autonomous car is trained in sunny weather scenes, but in some moments, the weather condition changes, then the detector will look at scenes that it is not trained to work with. In such cases, it may not perform at its optimal accuracy, or at different times in a day, the object detection performance can be different. Fig. 2 shows four example data with different light conditions in BDD100K dataset, from top-left to bottom-right: daytime, twilight, nighttime with street light, and nighttime without street light.

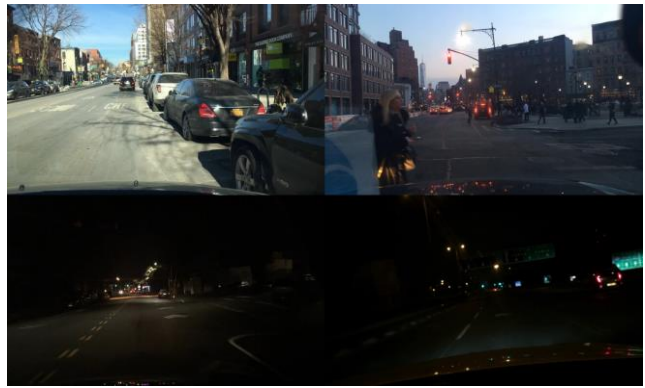

Fig. 2. Example data with different light conditions in BDD100K dataset.

For a superb object detector which can perform well in different contexts, a neural network architecture with sophisticated and heavy computational cost is desired for great accuracy of the system. However, deploying such bulky models on mobile devices for instant usage is undesirable because deployment to mobile devices has numerous stringent requirements on latency and computational resources. One way to solve this problem is to deploy the model to the cloud and call it whenever its service is required. However, this would require a high-quality internet connection which is not always available for moving objects like cars, thus it becomes a constraint in production. This problem can be solved if it is implemented in a smarter way; that is, different models are trained and each one is able to handle perfectly a specific situation. We then only need to call an alternative model sometimes when there is a significant change in the environmental condition. In this paper, we integrate an environmental condition classifier into our system. In order to send a warning signal to the cloud whenever there is a significant environmental change, the system merely requires an acceptable internet connection between the mobile device and the cloud.

## III. ROBUSTNESS ENHANCEMENT

This section provides two techniques integrated into a single system that can be applied to ADAS to improve the robustness of the object detection module.

### A. YOLO-based Object Detector

The core of an object detection module definitely is an object detection network. Generally, there are two genres of object detectors: one-stage object detectors which show high inference speed and considerable accuracy, and two-stage object detectors which yield higher object recognition and localization accuracy but with expensive computational cost and diminished speed [6]. Because of the superiority in speed, one-stage object detectors generally are integrated into real-time object detection systems and mobile devices such as mobile phones, autonomous vehicles, and drones. Since this research is in the direction of enhancing the robustness of object detectors in ADAS, one-stage detection scheme is chosen as the main type of network to investigate in this paper. In one-stage object detectors, an input image is passed through a backbone network to extract features and produce a final feature map with distilled information of objects, object recognition and localization of bounding boxes are performed directly on this feature map and done in a single pipeline without any other post-processing steps.

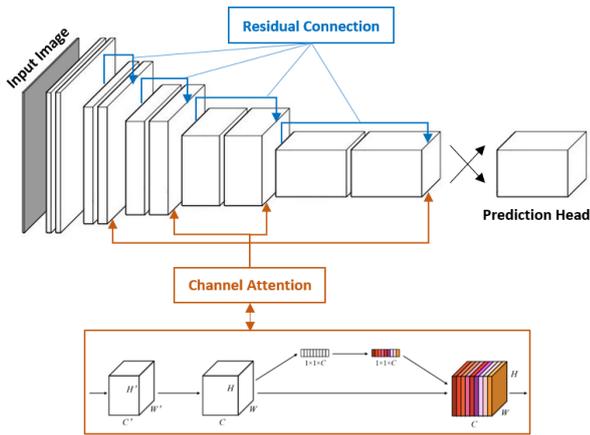

Fig. 3. Our proposed object detector (Backbone17-Det).

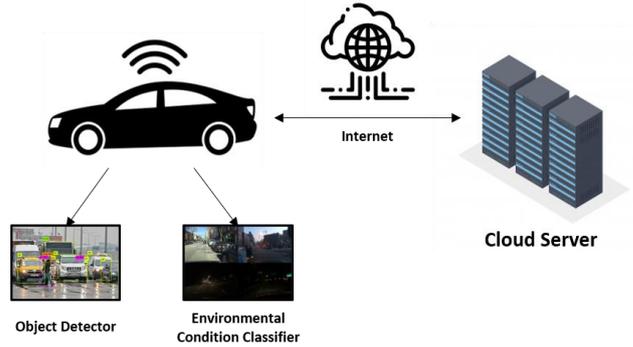

Fig. 4. Diagram of our system.

In this study, we propose an efficient one-stage object detector based on our prior work [6] whose architecture was constructed based on the classical object detection algorithm YOLO. Table 1 summarizes the backbone architectures of YOLOv2 [1] (Darknet-19) and our proposed object detector (Backbone17). In our prior work, we have performed experiments to verify the effectiveness of additional modules such as residual module [11], channel attention module (SE) [12], and spatial attention module (CBAM) [13] to choose the ones which are truly beneficial for improving our network performance. The experiments have shown that the network with SE outperforms the one with CBAM in our case [6]. Therefore, we integrate residual connection and SE into our network backbone architecture. Our network (Backbone17-Det) is visualized in Fig. 3, the residual connection is applied once in every level of feature map resolution while SE is applied before every downscaling step, specifically, SE is applied right after layer 5, layer 8, layer 13, and layer 17.

### B. Environmental Condition Classifier

During onboard deployment, the object detector of an autonomous driving car is expected to operate at a similar performance level as it is validated in training and testing stages even under varying and complex environmental conditions, but this seems infeasible. In fact, there are multiple factors that an onboard object detector relies directly and heavily on such as traffic density, road type, and time of the day, those factors impact directly the object detector and the instability and the degradation of its performance without warning is occasionally inevitable. This may lead the whole system to implement unsafe and risky actions due to unreliable object detection responses. The researchers in [14] introduce a cascaded neural network that monitors the performance of the object detector by predicting the quality of its mean average precision (mAP) on a sliding window of the input frames, the proposed cascaded network exploits the internal features from the deep neural network of the object detector. Similarly, but in a much simpler manner, we address this problem by proposing an environmental condition classifier to recognize and send a warning signal to the cloud whenever there is a significant environmental change. Because this system requires only an acceptable internet connection between the mobile device and the cloud, this connection protocol may help the authorities to take crucial actions from the station instead of leaving the car handle by themselves in the case the car needs human actions due to the significance of the situation. The environmental condition classifier is integrated along with the object detector in our system. Fig. 4 describes the diagram of our system.

### IV. EXPERIMENTS

In this section, we first provide experimental constraints, and the experimental results of the object detector as well as the environmental condition classifier are provided in order to prove the applicability and the scalability of our system. We also compare our result with other methods in terms of object detection efficiency.

TABLE I. DARKNET-19 AND BACKBONE17 ARCHITECTURES

| Layers | Types | Filters / Stride | | Output Resolution |
|---|---|---|---|---|
| | | Darknet-19 | Backbone17 | |
| 1 | Conv | 3 x 3 x 32 | 3 x 3 x 32 | 608 x 608 |
| | Maxpool | 2 x 2 / 2 | | 304 x 304 |
| 2 | Conv | 3 x 3 x 64 | 3 x 3 x 64 / 2 | 304 x 304 |
| | Maxpool | 2 x 2 / 2 | | 152 x 152 |
| 3 | Conv | 3 x 3 x 128 | 3 x 3 x 128 / 2 | 152 x 152 |
| 4 | Conv | 1 x 1 x 64 | 1 x 1 x 64 | 152 x 152 |
| 5 | Conv | 3 x 3 x 128 | 3 x 3 x 128 | 152 x 152 |
| | Maxpool | 2 x 2 / 2 | | 76 x 76 |
| 6 | Conv | 3 x 3 x 256 | 3 x 3 x 256 / 2 | 76 x 76 |
| 7 | Conv | 1 x 1 x 128 | 1 x 1 x 128 | 76 x 76 |
| 8 | Conv | 3 x 3 x 256 | 3 x 3 x 256 | 76 x 76 |
| | Maxpool | 2 x 2 / 2 | | 38 x 38 |
| 9 | Conv | 3 x 3 x 512 | 3 x 3 x 512 / 2 | 38 x 38 |
| 10 | Conv | 1 x 1 x 256 | 1 x 1 x 256 | 38 x 38 |
| 11 | Conv | 3 x 3 x 512 | 3 x 3 x 512 | 38 x 38 |
| 12 | Conv | 1 x 1 x 256 | 1 x 1 x 256 | 38 x 38 |
| 13 | Conv | 3 x 3 x 512 | 3 x 3 x 512 | 38 x 38 |
| | Maxpool | 2 x 2 / 2 | | 19 x 19 |
| 14 | Conv | 3 x 3 x 1024 | 3 x 3 x 1024 / 2 | 19 x 19 |
| 15 | Conv | 1 x 1 x 512 | 1 x 1 x 512 | 19 x 19 |
| 16 | Conv | 3 x 3 x 1024 | 3 x 3 x 1024 | 19 x 19 |
| 17 | Conv | 1 x 1 x 512 | 3 x 3 x 1024 | 19 x 19 |
| 18 | Conv | 3 x 3 x 1024 | | 19 x 19 |
| 19 | Conv | 1 x 1 x 1000 | | |

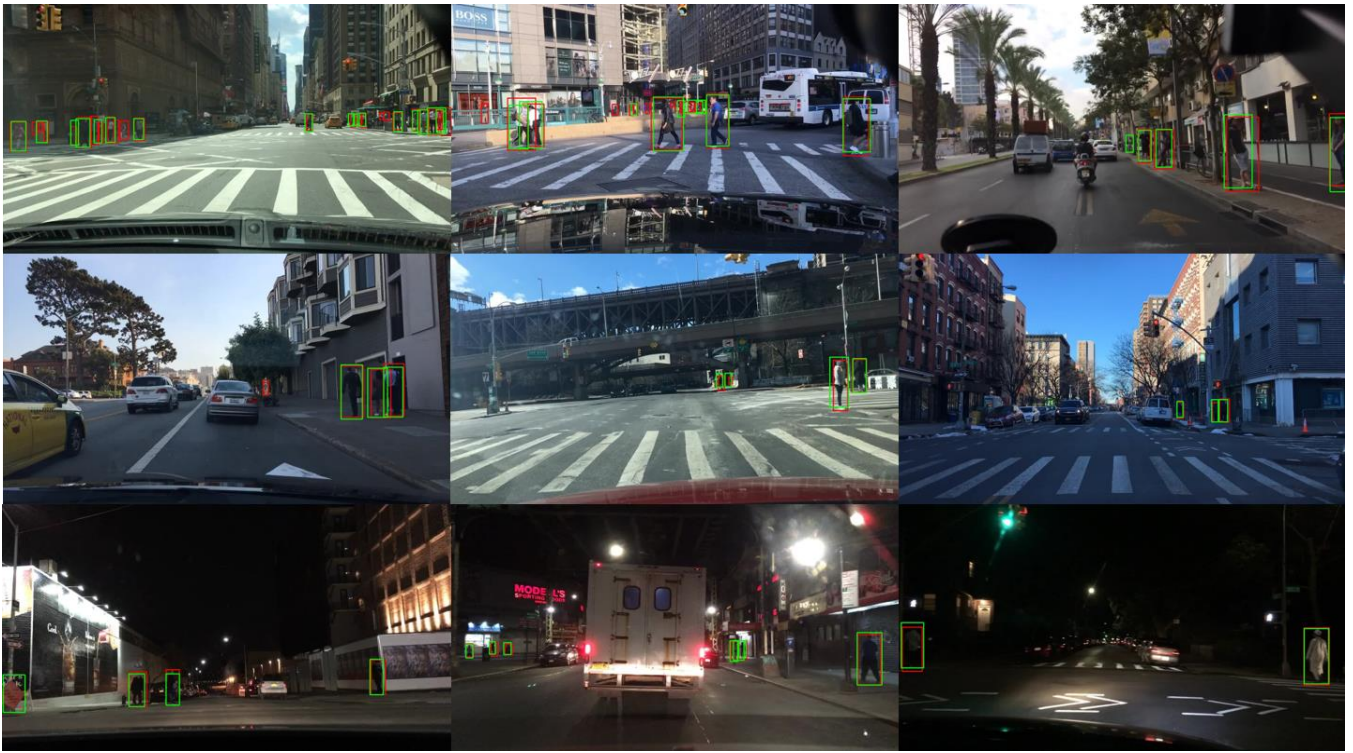

Fig. 5. Person detection results using our object detector (Backbone17-Det), red boxes indicate the ground truth, green boxes depict detection results.

*A. Experimental Constraints*

We propose the system in the scenario that we do not possess a real model of autonomous driving car. Therefore, the experiments are conducted in several constraints:

*1) Object detector:* Our object detector is not tested on a mobile device, but a PC. Hence, we only evaluate the accuracy of our object detector, not its inference speed, though we expect that our detector is able to obtain real-time inference speed on mobile devices due to its compactness.

*2) Environmental condition classifier:* Our system is validated on BDD100K dataset [9], this dataset does not supply image data in different weather conditions, we simulate the variation of surrounding environment contexts by using the change of light condition in different times of day as described in Fig. 2.

*B. Object Detector*

Our object detection network, Backbone17-Det, is trained on BDD100K [9]. BDD100K is a large-scale dataset with over 100K videos and its 2D bounding box annotations include 10 object categories: bus, traffic light, traffic sign, person, bike, truck, motor, car, train, and rider. For the sake of simplicity, we only concentrate on executing experiments and analysis on the person object category. The person data includes more than 22K images (91K person objects) for training and about 3.2K images (13K person objects) for testing. We train our network with multi-scale training strategy; that is, the input size scales from 320x320 to 608x608 during training and a specific resolution is chosen randomly for the input batch at the beginning of every iteration. The network is trained on 2 GeForce GTX TITAN X Graphics Cards, the batch size is 6 for training and is 2 for test. Training IoU threshold is 0.3. The total number of trained epochs is 120, the learning rate is ranged and dropped gradually from $10^{-4}$ at the first 2 epochs to $10^{-6}$ at the end of the training.

The proposed object detection scheme is evaluated in terms of Average Precision (AP) metric. We follow the PASCAL VOC convention by reporting AP at IoU = 0.5 ($AP_{50}$). Our network with input size 608x608 produces favorable detection outcomes of 43.6 AP corresponding to 7,978 detected objects out of 13,262 ground-truth objects with IoU = 0.5, and 60.4 AP corresponding to 10,608 detected objects with IoU = 0.3 ($AP_{30}$). In order to compare the results in [15] in terms of the $AP_{50}$ metric, two-stage object detection model Faster R-CNN [3] is trained and tested on BDD100K person data. The AP for our proposed scheme is 43.6 while Faster R-CNN shows 45.4. Note that the input image resolutions are 1000x600 and 608x608 for Faster R-CNN and our proposed scheme, respectively. Table II shows the performance comparison. Our person detection results are shown in Fig. 5, red boxes present the labels and the detection boxes are indicated in green.

*C. Environmental Condition Classifier*

Due to the experimental constraints explained above, we deploy a simple classifier that recognizes the light condition

TABLE II. OBJECT DETECTION PERFORMANCE COMPARISON

| Model | $AP_{30}$ (%) | $AP_{50}$ (%) | True Positives | Labels |
|---|---|---|---|---|
| Our | 60.4 | - | 10,608 (80.0%) | 13,262 |
| Our | - | 43.6 | 7,978 (60.2%) | 13,262 |
| Faster R-CNN | - | 45.4±5.2 | - | 13,262 |

```
Algorithm: Light Condition Classification
    Input: RGB image
    Output: light condition
    initialize threshold values {T_1, T_2, T_3}      // 0<T_1<T_2<T_3<255
    while system is running do
        convert RGB input image to grayscale
        compute average gray value (AGV)
        if AGV ≥ T_1 then
        light_condition ← daytime
        else if T_2 ≤ AGV < T_1 then
        light_condition ← twilight
        else if T_3 ≤ AGV < T_2 then
        light_condition ← nighttime with street light
        else
        light_condition ← nighttime without street light
        end if
        return light_condition
    end while
```

Fig. 6. Pseudocode of light condition classification algorithm.

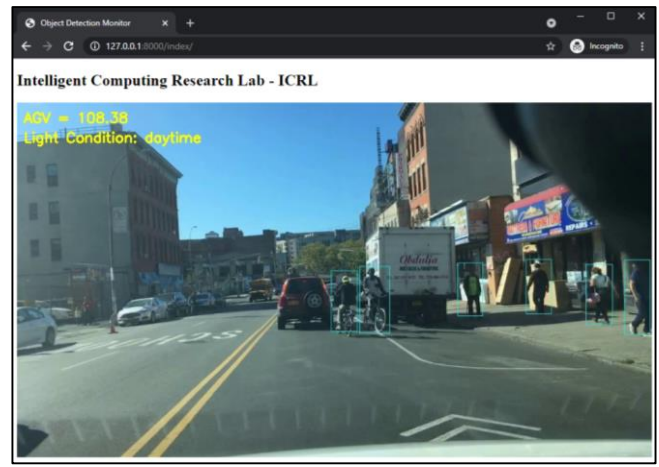

Fig. 7. Detection result is displayed on web browser.

of surrounding environment using an average gray value (AGV) of the grayscale of the RGB input image. We simulate our classifier with 4 light conditions as mentioned in Section II. We empirically choose 3 thresholds $T_1$, $T_2$, and $T_3$ of the AGVs to determine the light condition output, that is, if AGV $\geq T_1$, the captured scene is in daytime, if $T_2 \leq$ AGV $< T_1$, the captured scene is in twilight, if $T_3 \leq$ AGV $< T_2$, the captured scene is in nighttime with street light, otherwise, the captured scene is in nighttime without street light. We empirically choose $T_1 = 50$, $T_2 = 20$, and $T_3 = 10$. The pseudocode of the light condition classification algorithm is shown in Fig. 6.

The detected bounding boxes should be drawn on the input image for the sake of visualization and the result is displayed on a built-in monitor via a web browser. To this end, we create a web application based on our prior work [16] and establish a connection between the system and the cloud server for the purposes of surveillance and dealing with critical situations as explained throughout the paper. Therefore, our proposed system requires an acceptable internet connection. For the sake of efficiency, the system does not always send image data to the server, instead, object detection results are merely displayed on the built-in monitor and the system only sends a warning to the server whenever the classifier recognizes a significant change of the environmental condition, thus it does not require a high-quality internet connection, however, the better response speed, the safer the self-driving car. The web application with a simple interface is depicted in Fig. 7.

V. CONCLUSIONS AND FUTURE WORKS

In this work, we propose a novel enhancement for the robustness of the object detection module in advanced driver assistance systems (ADAS). A unified system integrating an efficient object detector and an environmental condition classifier is proposed. As object detectors have become an essential component in robotic systems, their performance during the deployment phase is critical to ensure the safety and reliability of the whole system. Since, however, modern self-driving cars are still so far from being applied widely due to their limitation of accuracy, more robust and stable ADAS should be developed. To this end, our system is proposed to improve the safety of a self-driving car itself and to enhance ADAS efficiency with reliable human supports from the server. The proposed system in this paper can be considered a potential for the scalability of the next generation of autonomous driving cars.

Although the experimental results have shown the contribution of the proposed system, there exist a room for improvement. Important aspects for future works include: (1) model quantization and knowledge distillation for enhancing the compactness of the object detection model and improving detection accuracy, and (2) validating the proposed system with other datasets with different environmental contexts.